\documentclass[10pt,twocolumn,letterpaper]{article}

\usepackage{iccv}
\usepackage{times}
\usepackage{epsfig}
\usepackage{graphicx}
\usepackage{amsmath}
\usepackage{amssymb}
\usepackage{multirow}

\usepackage[breaklinks=true,bookmarks=false]{hyperref}

\iccvfinalcopy % *** Uncomment this line for the final submission

\def\iccvPaperID{****} % *** Enter the ICCV Paper ID here

\ificcvfinal\fi

\begin{document}

%%%%%%%%% TITLE
\title{Technical Report: Disentangled Action Parsing Networks for Accurate Part-level Action Parsing}
\author{ Xuanhan Wang\textsuperscript{1} 
		\and Xiaojia Chen\textsuperscript{1} 
		\and Lianli Gao\textsuperscript{1} 
		\and Lechao Cheng\textsuperscript{2} 
		\and Jingkuan Song\textsuperscript{1}  
	\and \textsuperscript{1} Center for Future Media, University of Electronic Science and Technology of China, Chengdu, China
	\and \textsuperscript{2}Zhejiang Lab, Hangzhou, China \\
	{\tt\small wxuanhan@hotmail.com, josonchan1998@163.com, lianli.gao@uestc.edu.cn}\\
	{\tt\small chenglc@zhejianglab.com, jingkuan.song@gmail.com}
}

\maketitle
\ificcvfinal\thispagestyle{empty}\fi

%%%%%%%%% ABSTRACT
\begin{abstract}
   Part-level Action Parsing aims at part state parsing for boosting action recognition in videos. Despite of dramatic progresses in the area of video classification research, a severe problem faced by the community is that the detailed understanding of human actions is ignored. Our motivation is that parsing human actions needs to build models that focus on the specific problem.
   
   We present a simple yet effective approach, named disentangled action parsing (DAP). Specifically, we divided the part-level action parsing into three stages: 1) person detection, where a person detector is adopted to detect all persons from videos as well as performs instance-level action recognition; 2) Part parsing, where a part-parsing model is proposed to recognize human parts from detected person images; and 3) Action parsing, where a multi-modal action parsing network is used to parse action category conditioning on all detection results that are obtained from previous stages. With these three major models applied, our approach of DAP records a global mean of $0.605$ score in 2021 Kinetics-TPS Challenge.
\end{abstract}

%%%%%%%%% BODY TEXT
\section{Introduction}

The Part-level Action Parsing (PAP) task was firstly proposed as a part of the DeeperAction Challenge at ICCV 2021, which requires models to recognize a human action by compositional learning of body part state in videos. In this work, we present the disentangled action parsing (DAP), a modular framework for part-level action parsing research, which formed the basis for the submission to the Kinetics-TPS Challenge from \textbf{CFM-HAG} team.

The motivation behind the proposed method comes from the following observations: First, the process of part-level action parsing (PAP) is interpreted as a three-stage pipeline, which detects all person instances from all video frames, then solves the single-person part parsing problem, and finally predicts the action category conditioned on results that are obtained from previous stages. 
Secondly, the top-down strategy has become the dominant solution to various human-centric tasks, such as person detection \cite{DBLP:conf/cvpr/LinDGHHB17,maskrcnn}, keypoint detection \cite{pose_SunXLWang2019,pose_li2018crowdpose,pose_xiao2018simple} and human parsing \cite{densepose,densepose:parsingrcnn,densepose:ktn}. Generally, it divides a task into a sequence of sub-problems (e.g., detect-then-estimate), and orderly solves them by designing problem-specific neural networks (e.g., Faster-RCNN for detection and HRNet \cite{pose_SunXLWang2019} for estimation). 
Third, we find out that bottleneck of the PAP lies in the ``Part parsing'' stage for boosting the final performance. In particular, we conduct a diagnostic experiment by investigating a simple approach, where an extended FPN detector \cite{DBLP:conf/cvpr/LinDGHHB17} is adopted for instance-level recognition and a TPN \cite{yang2020tpn} model is used for video classification.
The experimental results are summarized in Tab.~\ref{tab.bottleneck}. From the results, we observe that the performance is dramatically improved when replacing the predictions of part with their ground truth. 
According to these observations, we believe that the precise human part-level parsing plays an important role for action parsing. In this technical report, we aim to study one problem: \textit{how to design an effective pipeline for supporting distinct sub-tasks (e.g., person detection, part parsing and action parsing), in particular improving the performance of \textbf{part parsing}.}

\begin{table}[h]
	\caption{Bottleneck analysis for part-level action parsing. `'$\checkmark$'' means the predictions are replaced by corresponding ground truth.}
	\resizebox{\linewidth}{!}{
		\begin{tabular}{|c|cc|c|c|}
			\hline
			\multicolumn{3}{|c|}{Frame-level}                                                                                            & \multicolumn{1}{c|}{Video-level}       & \multicolumn{1}{c|}{Metric}               \\ \hline
			\multicolumn{1}{|c|}{\multirow{2}{*}{Actor Detection}} & \multicolumn{2}{c|}{Part parsing}                                  & \multicolumn{1}{c|}{\multirow{2}{*}{Action parsing}} & \multicolumn{1}{c|}{\multirow{2}{*}{${Acc}^{p}$}} \\ \cline{2-3}
			\multicolumn{1}{|c|}{}                                  & \multicolumn{1}{c|}{part\_det} & \multicolumn{1}{c|}{state\_parsing} & \multicolumn{1}{c|}{}                  & \multicolumn{1}{c|}{}                  \\ \hline
			&                               &                                    &                                        &            33.32\%                  \\ \hline 
			\checkmark &                               &                                    &                                        &    35.33\%                           \\ \hline
						&              \checkmark                 &                                    &                                        &                 36.22\%             \\ \hline  
			&              \checkmark                 &           \checkmark                         &                                        &     72.46\%                          \\ \hline
			&                               &                                    &                             \checkmark           &              42.20\%                 \\ \hline
			&                               &                        \checkmark             &                                       &              45.60\%                 \\ \hline
			\checkmark &              \checkmark                 &                                    &                                        &      38.10\%                         \\ \hline 
			\checkmark &              \checkmark                 &           \checkmark                         &                                        &                 77.30\%               \\ \hline  
			\checkmark &                               &                                    &                  \checkmark                      &      44.76\%                         \\ \hline
			&              \checkmark                 &           \checkmark                         &          \checkmark                              &     93.41\%                          \\ \hline 
			\checkmark &              \checkmark                 &           \checkmark                         &            \checkmark                            &                 99.90\%               \\ \hline
		\end{tabular}
	}
	\label{tab.bottleneck}
\end{table}

\section{Approach}
\begin{figure*}[ht]
	\begin{center}
		\includegraphics[width=0.9\linewidth]{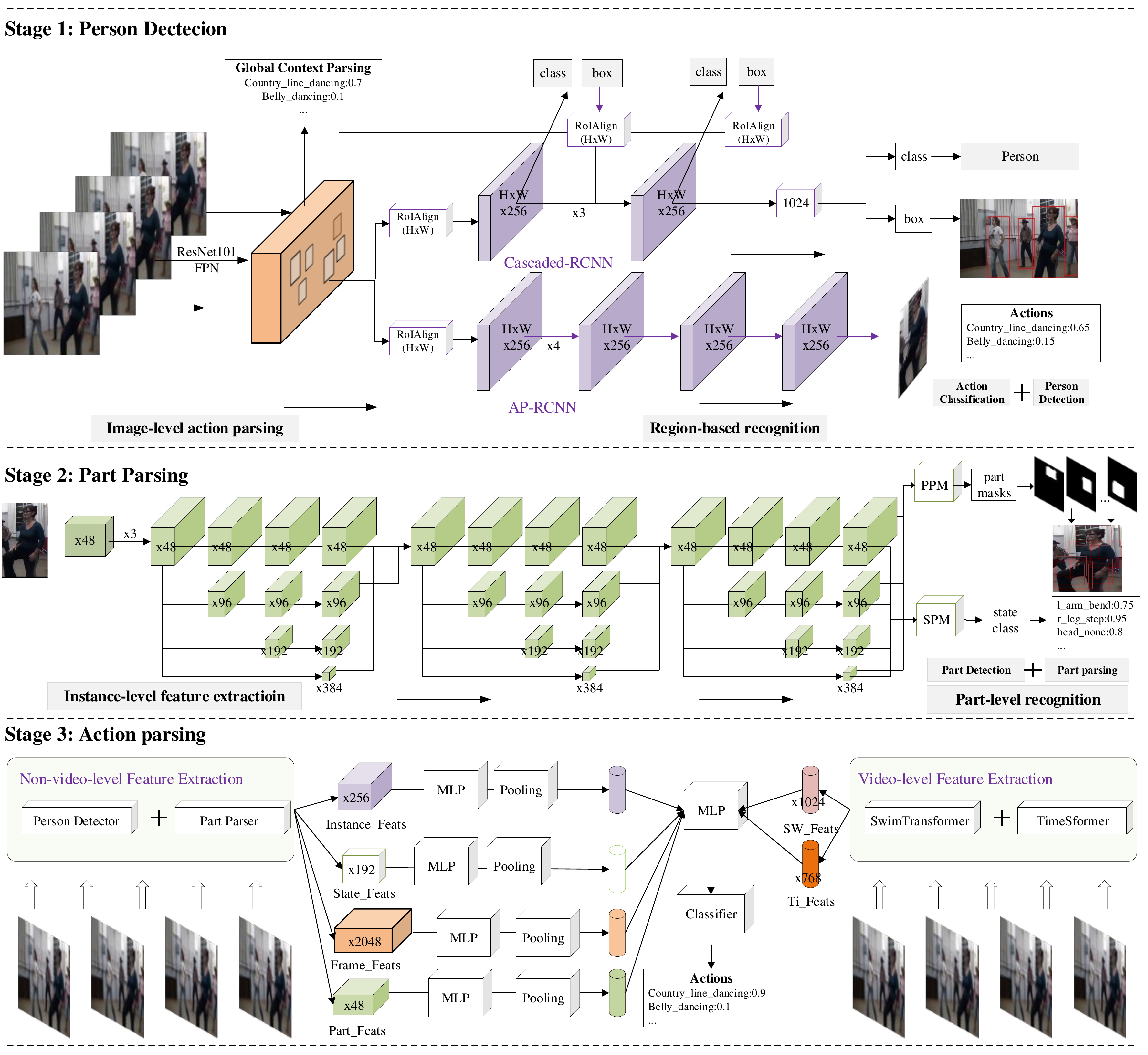}
	\end{center}
	\caption{The overview of the proposed pipeline for part-level action parsing.}
	\label{fig:framework}
\end{figure*}

The overall pipeline of the proposed method is shown in Fig.~\ref{fig:framework}. We follow the design rule of ``divide-and-conquer'' and build our PAP model by designing three task-specific neural networks. Next, we orderly present the proposed models for person detection, part parsing and action parsing. 

\subsection{Person Detection}
To detect persons from video frames, we follow previous works and adopt the state-of-the-art two-stage object detection approach based on Cascaded-RCNN \cite{Cai_2019}, which starts by region proposal generation, and then refines each proposal in a cascaded manner for predicting objects' locations with their categories. In addition, human actions can be inferred either from person instances or visual context. Thus, we further extend the Cascaded-RCNN by adding one RCNN branch for instance-level action parsing and a global context module for frame-level action parsing.

\noindent\textbf{Instantiation:} To implement our person detector, we take the Cascaded-RCNN with ResNet-101 as the basic detection model. In parallel with the person detection branch, two sub-networks are built, involving the Action Parsing RCNN (AP-RCNN) and the Global Context Parsing (GCP). Specifically, the AP-RCNN is constructed by four consecutive convolutional layers with 256 channels followed by a linear layer with $C$ channels, where convolutional layers output the instance-level action features $f_{ia} \in \mathbb{R}^{256}$ and $C$ denotes the number of action categories. As for the GCP, we simple adopt a global average pooling layer followed by one fully connected layers, where the first layer outputs frame-level action features $f_{c} \in \mathbb{R}^{2048}$ with 2048 dimensions and the second one is the $C$-way classifier. As a result, the person detector outputs a set of predictions, including the bounding box $\{\mathcal{P}_{cls} \in \mathbb{R}^{2}, \mathcal{P}_{box} \in \mathbb{R}^{8}\}$, the instance-level categorical distribution $\mathcal{C}_{ins} \in \mathbb{R}^{C}$ and the frame-level categorical distribution $\mathcal{C}_{img} \in \mathbb{R}^{C}$.

\noindent\textbf{Learning objectives:} To enable the model to perform person detection, we design our learning objectives as follows:
\begin{equation}
\begin{array}{lll}
\ell_{cls} &=  Cross\_Entropy(\mathcal{P}_{cls}, \mathcal{P}_{cls}^{*}) & \\
\ell_{box} &=  SmoothL1(\mathcal{P}_{box}, \mathcal{P}_{box}^{*}) & \\
\ell_{ins} &=  Cross\_Entropy(\mathcal{C}_{ins}, \mathcal{C}_{ins}^{*}) & \\
\ell_{img} &=  Cross\_Entropy(\mathcal{C}_{img}, \mathcal{C}_{img}^{*}) & \\
\ell_{det} &=  \ell_{cls} + \ell_{box} + \ell_{ins} + \ell_{img} & \\
\end{array}
\label{equ.detector_loss} 
\end{equation}
where $\mathcal{P}_{cls}^{*}$,$\mathcal{P}_{box}^{*}$,$\mathcal{C}_{ins}^{*}$ and $\mathcal{C}_{img}^{*}$ is the corresponding ground truth.
\subsection{Part Parsing}
In this section, we introduce our method to build part parsing network for facilitating part-level recognition. Given a person image $\mathit{I} \in \mathbb{R}^{H \times W \times 3}$ of size $H \times W$, the goal of part-level recognition is to predict the location of each body part with action states, involving two sub-problems (i.e., part detection and state classification). In addition, the HRNet \cite{pose_SunXLWang2019} has shown superiority on many human-centric visual tasks. Therefore, we build our model on the top of HRNet and extend it to the Parsing-HRNet for part-level recognition. The extended version consists of three major components: a visual feature network for visual feature generation, the part parsing modular (PPM) for part detection and a state parsing modular (SPM) for state recognition. 

\noindent\textbf{Instantiation:} Our part parsing model is built on the top of HRNet-W48 \cite{pose_SunXLWang2019}. Formally, let $\mathbf{f} \in \mathbb{R}^{d \times H_h \times W_h}$ be the visual features extracted from the visual feature network (i.e., HRNet-W48, $d=48$). Then, we separately feed visual features into the PPM for part detection and the SPM for state recognition. The PPM is constructed by three convolution layers, where the first two layers are used to generate part features $f_{pa} \in \mathbb{R}^{d \times H_h \times W_h}$, and the third one is the output layer that predicts $K$ heatmaps $\mathcal{O} \in \mathbb{R}^{K \times H_h \times W_h}$ for part detection. Furthermore, the SPM first applies a global average pooling on visual features for context feature generation. Then the context features are transformed into state features $f_{sta} \in \mathbb{R}^{192}$ by using two fully connected layers. Conditioning on the state features, the SPM outputs $S$-way probability distribution $\mathcal{D}_{sta} \in \mathbb{R}^{S}$ by using one linear layer, where $S$ is the number of action state classes.

\noindent\textbf{Learning objectives:} Two loss functions are used to enable the proposed part parsing model to perform part-level parsing. For part heatmap regression, we use the mean square error as the learning objective, and we use the standard focal loss \cite{8417976} for state classification. The overall loss function is defined in Eq.~\ref{equ.parsing_loss}:
\begin{equation}
\begin{array}{lll}
\ell_{p} &=  MSE(\mathcal{O}, \mathcal{O}^{*}) & \\
\ell_{s} &=  FocalLoss(\mathcal{D}_{sta}, \mathcal{D}_{sta}^{*}) & \\
\ell_{part} &=  \ell_{p} + \lambda\ell_{s} & \\
\end{array}
\label{equ.parsing_loss} 
\end{equation}
where $\lambda$ is the hyperparameter and it is set to 0.5 for balancing training. $\mathcal{O}^{*}$ is the ground-truth heatmaps and $\mathcal{D}_{sta}^{*}$ is the ground-truth action state. Note that each heatmap is defined as a part-specific binary mask, where each pixel across part area is set to 1, and 0 otherwise. 
\subsection{Action Parsing}
Video-level action feature is the key to precisely recognize human actions from videos. In this work, video-level action features are generated from multi-modal features, including non-video-level features and visual video-level features. In particular, non-video-level features involve frame-level action features, instance-level features, part features and part state features. After feature extraction, multiple MLP networks are used  for final action recognition. 

\noindent\textbf{Feature extraction:} For non-video-level feature extraction, we directly extract multiple features from the latent layers in the person detection network or the part parsing network, since these two models are fully trained in previous stages.  Specifically, we feed each video frame into person detection network and extract outputs from latent layers, including frame-level features $f_{c}$ from GCP module and multiple instance-level action features $\{f_{ia}\}$ from AP-RCNN. As for part-level features, we feed each person image into part parsing network and extract features from two sub-modules, involving spatially pooled part features $f_{pa}^{'} \in \mathbb{R}^{d}$ from PPM and action state features $f_{sta}$ from SPM. Formally, non-video-level features for each video are orderly denoted as: 1) frame-level action features $\hat{f_{c}} \in \mathbb{R}^{T \times 2048}$; 2) instance-level action features $\hat{f_{ia}} \in \mathbb{R}^{T \times P \times 256}$; 3) part features  $\hat{f_{pa}^{'}} \in \mathbb{R}^{T \times P \times 48}$ and 4) state features $\hat{f_{sta}} \in \mathbb{R}^{T \times P \times 192}$, where $T$ denotes the number of sampled frames and $P$ is the number of person boxes predicted in each frame. In our implementation, 32 frames are randomly sampled from each video. For instance-level feature extraction, top-10 person boxes with the highest score in each frame are considered. 
Besides, we also extract video-level features from two trained action models, including TimeSformer \cite{bertasius2021spacetime} and SwimTransformer \cite{liu2021video}. Formally, we denote the $f_{v}^{t} \in \mathbb{R}^{768}$ is the video-level feature extracted from TimeSformer, and $f_{v}^{s} \in \mathbb{R}^{1024}$ is the one from SwimTransformer.

\noindent\textbf{Inference:} Given non-video-level features, we first feed them into a MLP network, where it contains two fully-connected layers. Then we orderly perform max-pooling along with $P$ dimension and $T$ dimension, resulting in video-level features. Finally, we concatenate all video-level features together, and feed this feature into a linear layer to get action prediction.

\section{Experiments}
In this section, evaluation datasets and our implementation details are introduced at first. Then, we conduct detailed ablation studies to investigate the variants of each task-specific model. Finally, we provide our final version that forms the basis for the submission to the Kinetics-TPS Challenge.

\subsection{Experimental Settings}
We evaluate our models on the $2021$ KineticsTPS dataset. There are 3809 annotated videos in training set. Since the test set only provides video-level annotations, we randomly pick 30\% of training set as the minival set in the validation phase, resulting in 2686 videos for training and 1123 videos for validation. In the test phase, all models are trained in whole training set with 3809 videos and tested on the official server\footnote{\url{https://competitions.codalab.org/competitions/32360}}. We adopt the Average Precision ($mAP$), mean class accuracy ($Acc$) as well as video accuracy conditioning on Part State Correctness (${Acc}^{p}$) as the evaluation metrics. Our models are implemented based on OpenMMLab\footnote{\url{https://github.com/open-mmlab}} on an Ubuntu server with eight Tesla V100 graphic cards. For optimization, the details of training configuration is summarized in Tab.~\ref{tab.config}. 

\begin{table}[ht]
	\centering
	\caption{Training configurations.}
	\renewcommand\arraystretch{1.6}
	\huge 
	\resizebox{\linewidth}{!}{
		\begin{tabular}{|c|c|c|c|c|c|}
			\hline
			Model           & Example model            & Training epochs & Learning rate & Drop steps       & Solver \\ \hline
			Person detection & Cascaded-RCNN(ResNet101) & 12              & 0.02          & Linaer\{8,11\}     & SGD    \\ \hline
			Part Parsing    & HRNet-W48                & 40              & 0.0001        & Linear\{30,35\}    & Adam   \\ \hline
			Action Parsing  & SwimTransforer-Base      & 30              & 0.001         & ConsineAnnealing & AdamW  \\ \hline
		\end{tabular}
	}
	\label{tab.config}
\end{table}

\subsection{Ablation Study}

In this section, we investigate the variants of proposed disentangled action parsing (DAP) method. 

\noindent\textbf{Person detection.} We choose the Cascaded-RCNN with ResNet101 as the baseline model. Then we investigate the effect of the GCP module and the AP-RCNN by gradually incorporating them into baseline mode. The experimental results are summarized in Tab.~\ref{tab.actor_det}. From the results, we observe that the GCP and AP-RCNN bring minor effects to the AP score.   

\begin{table}[ht]
	\centering
	\caption{Ablation study. The investigation of the person detector variants.}
	\resizebox{0.75\linewidth}{!}{
	\begin{tabular}{|ccc|c|}
		\hline
		Cascaded-RCNN & GCP & AP-RCNN & mAP \\ \hline
		\checkmark	&           &          &  76.6\%   \\ \hline
		\checkmark	&        \checkmark     &        &  77.0\%   \\ \hline
		\checkmark	&        \checkmark     &  \checkmark      &  76.0\%   \\ \hline
	\end{tabular}
	}
	\label{tab.actor_det}
\end{table}

\begin{table*}[ht]
	\centering
	\caption{Ablation study. The investigation of part parsing variants.}
	\begin{tabular}{|ccccc|c|}
		\hline
		Shared-heatmaps & Part-heatmaps & State-heatmaps & State-vectors & Focal loss & ${Acc}^{p}$ \\ \hline
		\checkmark	&               &                &               &            &     50.9\%   \\ \hline
		&     \checkmark          &       \checkmark         &               &                &  51.3\%   \\ \hline
		&     \checkmark          &                &     \checkmark          &              &  51.7\%   \\ \hline
		&     \checkmark        &                &       \checkmark        &    \checkmark       &  53.2\%   \\ \hline
	\end{tabular}
	\label{tab.parsing_net}
\end{table*}

\begin{table}[ht]
	\centering
	\caption{Ablation study. The effect of video features on minival set.}
	\renewcommand\arraystretch{1.5}
	\resizebox{\linewidth}{!}{
		\begin{tabular}{|c|ccc|cc|cc|}
			\hline
			\multicolumn{1}{|c|}{Frame-level} & \multicolumn{3}{c|}{Instance-level} & \multicolumn{2}{c|}{Video-level} & \multicolumn{2}{c|}{Metric} \\ \hline
			$\hat{f_{c}}$ & $\hat{f_{ia}}$ & $\hat{f_{pa}^{'}}$ & $\hat{f_{sta}}$ & $f_{v}^{t}$ & $f_{v}^{s}$ & $Acc$ & ${Acc}^{p}$\\ \hline
			\checkmark	&                 &               &             &  &  &    73.0\%   & 46.1\%   \\ \hline
			&          \checkmark       &        \checkmark       &   \checkmark          &  &  &      74.2\%  & 46.4\%  \\ \hline
			&                &               &  & \checkmark &  &     85.6\%   &    53.2\%      \\ \hline
			&                &               &  &  & \checkmark &     87.0\%   &    53.6\%      \\ \hline \hline
			\multicolumn{8}{|c|}{Model Ensemble} \\ \hline
			\multicolumn{1}{|c|}{Frame-level} & \multicolumn{3}{c|}{Instance-level} & \multicolumn{2}{c|}{Video-level} & $Acc$ & ${Acc}^{p}$ \\ \hline
			\checkmark	&     \multicolumn{3}{c|}{\checkmark}         &   &  & 79.3\% & 49.7\%       \\ \hline
			\checkmark	&     \multicolumn{3}{c|}{\checkmark}         &  \multicolumn{2}{c|}{\checkmark} & 93.2\%   &  57.5\%    \\ \hline
		\end{tabular}
	}
	\label{tab.action_cls_val}
\end{table}

\begin{table}[ht]
	\centering
	\caption{Ablation study. The effect of video features on test set.}
	\renewcommand\arraystretch{1.5}
	\resizebox{\linewidth}{!}{
		\begin{tabular}{|c|ccc|cc|c|}
			\hline
			\multicolumn{1}{|c|}{Frame-level} & \multicolumn{3}{c|}{Instance-level} & \multicolumn{2}{c|}{Video-level} & Metric \\ \hline
			$\hat{f_{c}}$ & $\hat{f_{ia}}$ & $\hat{f_{pa}^{'}}$ & $\hat{f_{sta}}$ & $f_{v}^{t}$ & $f_{v}^{s}$ & $Acc$ \\ \hline
			\checkmark	&                 &               &             &  &  &    73.8\%      \\ \hline
			&       \checkmark          &  \checkmark              &   \checkmark              &  &  &   79.0\%     \\ \hline
			&                &               &  & \checkmark &  &     85.0\%             \\ \hline
			&                &               &  &  & \checkmark &     87.5\%             \\ \hline \hline
			\multicolumn{7}{|c|}{Model Ensemble} \\ \hline 
			\multicolumn{1}{|c|}{Frame-level} & \multicolumn{3}{c|}{Instance-level} & \multicolumn{2}{c|}{Video-level} & $Acc$ \\ \hline
			\checkmark	&     \multicolumn{3}{c|}{\checkmark}         &   &  & 82.4\%         \\ \hline
			\checkmark	&       \multicolumn{3}{c|}{\checkmark}        & \multicolumn{2}{c|}{\checkmark} & 94.1\%         \\ \hline
		\end{tabular}
		\label{tab.action_cls_test}
	}
\end{table}
\begin{table}[ht]
	\centering
	\caption{Kinetics-TPS Challenge results on \textit{test} set.}
	\resizebox{0.75\linewidth}{!}{
	\begin{tabular}{|c|c|c|}
		\hline
	Ranks &	Method          & ${Acc}^{p}$                 \\ \hline
	(1) &	yuzheming       & 0.630532           \\ \hline
	(2)	&Sheldong        & 0.613722           \\ \hline
	(3)	&\textbf{Ours}   & \textbf{0.605059 } \\ \hline
	(4) 	&fangwudi        & 0.590167          \\ \hline
	(5)	&uestc.wxh       & 0.536067           \\ \hline
	(6)	&hubincsu        & 0.490984           \\ \hline
	(6) 	&scc1997         & 0.490984          \\ \hline
	(7)	&KGH             & 0.486483           \\ \hline
	(8)	&zhao\_THU        & 0.434311           \\ \hline
	(9)	&TerminusBazinga & 0.396735           \\ \hline
	(10)	&cjx\_AILab       & 0.370753          \\ \hline
	(11)	&xubocheng       & 0.358834          \\ \hline
	(12)	&haifwu          & 0.247614          \\ \hline
	(13)	&Aicity          & 0.189669          \\ \hline
	(14)	&fog             & 0.188455          \\ \hline
	\end{tabular}
	}
	\label{tab.challenge_test}
\end{table}
\noindent\textbf{Part parsing.} In this section, we compare four settings: 1) Shared heatmaps, where the part parsing network outputs action state heatmaps which can be decoded to part location and action states. 2) Separated heatmaps, where the part parsing model outputs part heatmaps and state heatmaps for part decoding and state decoding, respectively. 3) The part parsing model outputs part heatmaps but predicts state onehot vector. 4) Focal loss that used for state classification during training. To evaluate these models, we use the TimeSformer as the video-level classification model that has video accuracy of 85.6\% on minival set. We fix the video-level classification model throughout above experiments. The experimental results are summarized in Tab.\ref{tab.parsing_net}. From the results, we observe that separated version is better than shared version in terms of ${Acc}^{p}$ score. The parsing model can be further improved when applying focal loss. This is because the annotation of state classes exists characteristic of imbalance, in particular the number of class 'none' is much higher than that of others. 

\noindent\textbf{Action parsing.} In this section, we investigate the effect of action features. The experimental results are summarized in Tab.~\ref{tab.action_cls_val}. From the results, we observe that the multiple features are critical. With non-video-level features, the accuracy score is $\sim$74\%, which implies human actions can be well inferred from person states or frame-level context. In addition, ensemble multiple predictions as the final action scores can bring significant improvement. For example, the video-level accuracy can be increased to 93.2\% when ensemble all predictions. With the improved accuracy, the ${Acc}^{p}$ score has been increased to 57.5\% which is the best performance so far. Similar performance can be observed in Tab.\ref{tab.action_cls_test}.

\noindent\textbf{Predictions on test set.} With above experiences, we build the final version of DAP and apply it to obtain the predictions results on test set. Specifically, we use the extended Cascaded-RCNN trained on all non-split training set as the person detector, and we apply it to predict human boxes. With predicted human boxes, we choose the best variants of HRNet-W48 that explored in Tab.\ref{tab.parsing_net} and apply it to predict part locations and part states. As for action parsing, we ensemble multiple predicted results as the final action scores which has a video-level accuracy of 94.1\% on test set. With those predicted results, we have formed the submission to Kinetics-TPS challenge. The Tab.~\ref{tab.challenge_test} summaries the methods from the leaderboard\footnote{\url{https://competitions.codalab.org/competitions/32360\#results}} of Kinetics-TPS Challenge. From the results, we see that the proposed DPA obtains 60.5\% in ${Acc}^{p}$, which ranks third place on the test set.

\section{Conclusion}
In this report, we present a simple yet effective approach for part-level action parsing, named disentangled action parsing (DAP). We divide the part-level action parsing into three stages, involving person detection, part parsing and action parsing. Following this, we design three models, including the cascaded-rcnn variant for person detection, the variant of HRNet for part parsing and multi-modal action parsing model for video-level action parsing. Extensive experiments on Kinetics-TPS dataset demonstrate the effectiveness of the proposed method, and it obtain 60.5\% ${Acc}^{p}$ score in 2021 Kinetics-TPS Challenge.
{\small
\bibliographystyle{ieee_fullname}
\bibliography{egbib}
}

\end{document}